%% file: main.tex
\documentclass[conference,letterpaper]{IEEEtran}
\usepackage[T1]{fontenc}
\usepackage[utf8]{inputenc}
\usepackage{cite}
\usepackage{amsmath,amssymb,bm}
\usepackage{graphicx}
\usepackage{booktabs,array,tabularx}
\usepackage{microtype}
\usepackage{tikz}
\usetikzlibrary{arrows.meta,positioning,calc}
\usepackage{url}
\usepackage[hidelinks]{hyperref}
\hypersetup{
  pdftitle={Knowledge-Graph-Augmented Chronos-2 for HEC-RAS Surrogate Forecasting},
  pdfauthor={Edward Holmberg; Elias Ioup; Mahdi Abdelguerfi},
  pdfsubject={Prepared for submission to the 6th Workshop on Knowledge Graphs and Big Data (KGBigdata2026), IEEE BigData 2026},
  pdfkeywords={knowledge graphs, surrogate modeling, HEC-RAS, Chronos-2, hydraulic forecasting, time-series foundation models},
  pdflang={en-US}
}
\graphicspath{{figures/}}
\newcommand{\R}{\mathbb{R}}
\newcommand{\clip}{\operatorname{clip}}

\newcommand{\SSE}{\operatorname{SSE}}

\newcommand{\argmax}{\operatorname*{arg\,max}}
\newcolumntype{L}[1]{>{\raggedright\arraybackslash}p{#1}}
\newcolumntype{Y}{>{\raggedright\arraybackslash}X}
\title{Knowledge-Graph-Augmented Chronos-2 for HEC-RAS Surrogate Forecasting}
\input{heading_punctuation}

\author{\IEEEauthorblockN{Edward Holmberg}
\IEEEauthorblockA{\textit{Canizaro-Livingston Gulf States} \\
\textit{Center for Environmental Informatics}\\
New Orleans, LA, USA \\
eholmber@uno.edu}
\and
\IEEEauthorblockN{Elias Ioup}
\IEEEauthorblockA{\textit{Center for Geospatial Sciences} \\
\textit{Naval Research Laboratory}\\
Stennis Space Center, Mississippi, USA \\
elias.z.ioup.civ@us.navy.mil}
\and
\IEEEauthorblockN{Mahdi Abdelguerfi}
\IEEEauthorblockA{\textit{Canizaro-Livingston Gulf States} \\
\textit{Center for Environmental Informatics}\\
New Orleans, LA, USA \\
gulfsceidirector@uno.edu}
}
\begin{document}
\bstctlcite{IEEE:BSTcontrol}
\maketitle

\begin{abstract}
We investigate whether coupling a time-series foundation model to hydraulic project knowledge improves surrogate forecasting of HEC-RAS water-surface elevation (WSE). We present KG-Chronos-2, which combines a frozen Chronos-2 predictor with exact-state residual decoding, graph-conditioned historical retrieval, and input-aligned correction. We compare the method with persistence, a residual LSTM, project-conditioned recurrent GeoFNO, a hydraulic DCRNN-style model, and frozen Chronos-2. Task-specific fitting uses the 2008 simulation. Evaluation covers 64 fixed 24-hour windows from the 2011 and 2002 simulations at 4,675 cross sections in 71 reaches on a shared geometry. KG-Chronos-2 achieves event-balanced root-mean-square error 0.246970 in native WSE units. It reduces RMSE by 14.13\% relative to frozen Chronos-2, 29.38\% relative to the hydraulic DCRNN-style model, and 39.54\% relative to recurrent GeoFNO. The 95\% hierarchical-bootstrap interval for its event-balanced RMSE difference from frozen Chronos-2 is $[-0.075177,-0.016317]$. KG-Chronos-2 also achieves the lowest active-window and final-lead RMSE among the six completed systems. These results support coupling a frozen temporal predictor to project knowledge for warm-start HEC-RAS forecasting on the fixed benchmark.
\end{abstract}


\section{Introduction}\label{sec:intro}
\textbf{Motivation.} The Hydrologic Engineering Center's River Analysis System (HEC-RAS) computes river hydraulics from project geometry, cross sections, structures, and boundary conditions~\cite{hecras}. The computational cost of numerical time stepping motivates data-driven surrogates for near-term forecasting. Recent neural-operator approaches use these solved trajectories to train forecasting surrogates~\cite{holmberg2025}. We now investigate a foundation-model approach to the same objective: predict the next 24 hours of water-surface elevation (WSE) from an available history, bypassing numerical time stepping over the forecast horizon.

\textbf{Research question.} Can a frozen time-series foundation model coupled to hydraulic knowledge improve HEC-RAS surrogate forecasting relative to classical, locally trained temporal, spatial, and graph-recurrent approaches? Chronos-2 supplies a pretrained multivariate predictor~\cite{chronos2}. Native project records supply cross-section membership, reach relationships, hydraulic attributes, and prescribed inputs. We connect these records to the predictor through numerical retrieval and residual adaptation.

\textbf{Approach.} KG-Chronos-2 preserves the exact current WSE field and forecasts low-rank increments with frozen Chronos-2. A project graph conditions the selection of historical response windows and aligns prescribed inputs with receiving reaches. A gated adapter then corrects the retrieval-fused trajectory. The method retains the pretrained temporal weights while fitting its project-specific representation and correction parameters on the source simulation.

\textbf{Benchmark.} We evaluate six completed systems on identical cross sections, forecast origins, and targets: persistence, residual LSTM, project-conditioned recurrent GeoFNO, hydraulic DCRNN-style, frozen Chronos-2, and KG-Chronos-2. The comparisons address two questions. The broader benchmark measures competitiveness across forecasting approaches. The shared-backbone comparison measures the added value of the complete knowledge-coupling layer. We average the LSTM and graph-recurrent results over three training seeds.

\textbf{Contribution and result.} We connect native hydraulic project knowledge to a frozen temporal predictor and evaluate the resulting surrogate. KG-Chronos-2 reduces event-balanced RMSE by 14.13\% relative to frozen Chronos-2 and by 29.38\% relative to the hydraulic DCRNN-style model, the strongest locally trained comparator by mean RMSE. A direct paired bootstrap interval favors the coupled system over frozen Chronos-2 on the recorded two-event benchmark.

\section{Background}\label{sec:background}
\subsection{HEC-RAS Surrogate Forecasting}
Let $\bm y_t\in\R^N$ denote WSE at $N$ cross sections, $\bm u_t$ denote prescribed project inputs, and $\mathcal G$ denote the project knowledge representation. At origin $t$, a surrogate maps the available state history, project records, and permitted input schedule to $\widehat{\bm y}_{t+1:t+H}$. In this study, $N=4{,}675$, $H=24$, and the output cadence is hourly. Each forecast produces a $24\times4{,}675$ array of future WSE values.

We call this task \emph{warm-start state continuation}. The benchmark supplies the exact simulation-derived history and current field. The project-conditioned models also receive prescribed input schedules through the forecast horizon. Future HEC-RAS output fields serve exclusively as evaluation references. Forecast error measures agreement with the numerical solver's cross-section WSE. 

\subsection{Hydraulic Knowledge for Numerical Forecasts}
Knowledge graphs connect entities through relations and attributes~\cite{hogan2021}. Our implementation combines a numerical reach graph with linked catalogs of cross sections, hydraulic descriptors, input channels, and historical response windows. Cross-section membership localizes the WSE field; reach relationships supply neighborhood context; and input mappings associate prescribed disturbances with receiving reaches. These records guide numerical retrieval and residual correction around the Chronos forecast.

\section{Related Work}\label{sec:related}
\subsection{Hydraulic and Recurrent Surrogate Models}
Fourier neural operators (FNOs) learn field mappings through spectral transformations~\cite{fno}. Geo-FNO extends this formulation to general geometries through learned deformations~\cite{geofno}, and gated recurrent units (GRUs) encode temporal history~\cite{gru}. Recent work combines recurrent and Fourier-operator components with native hydraulic project inputs~\cite{holmberg2025}.

We train the recurrent comparator under the present benchmark. Its \emph{recurrent GeoFNO} implementation combines a GRU with one-dimensional spectral blocks along index-ordered cross sections and shares weights across reaches. This newly fitted configuration represents the published model family.

Zoch \emph{et al.} incorporate the Saint-Venant equations into physics-informed neural networks for single-river stage prediction~\cite{zoch2025pinn}. Holmberg \emph{et al.} introduce the Learned Response-Field Inertia Operator (LRFIO) for native-cell HEC-RAS 2D WSE prediction~\cite{holmberg2026lrfio}. LRFIO learns an increment-response operator and deploys it through closed-form rollout. These studies provide equation-informed and increment-based approaches to solver emulation. Our method couples a frozen temporal model to project-conditioned historical retrieval and residual correction.

Long short-term memory (LSTM) networks use gated recurrent state to learn sequence relationships~\cite{hochreiter1997lstm}. Diffusion convolutional recurrent neural networks (DCRNNs) combine graph diffusion with recurrent encoder--decoder forecasting~\cite{li2018dcrnn}. We evaluate a residual LSTM on the same latent increments as Chronos and a hydraulic DCRNN-style adaptation on all original cross sections. These comparators test locally learned temporal and graph-recurrent forecasting under the common output protocol.

\subsection{Geospatial Data and Hydraulic Representation}
Tu \emph{et al.} address interoperability among heterogeneous commercial geographic information systems~\cite{tu2002interoperability}. Chung \emph{et al.} describe the Geospatial Information Distribution System (GIDS) in an object-database setting~\cite{chung2001gids}. Wilson \emph{et al.} use XML-enabled technology for geographical data interchange in GIDB~\cite{wilson2003gidb}. These studies address the integration and exchange of geospatial records. Our project interface links native identifiers across WSE arrays, reach membership, geometry attributes, and prescribed inputs.

Abdelguerfi's edited volume examines 3D geospatial reconstruction and interoperability for synthetic environments~\cite{abdelguerfi2001reconstruction}. Flanagin \emph{et al.} develop hydraulic splines for modeling river-channel geometries~\cite{flanagin2007splines}. Ladner \emph{et al.} bring together spatiotemporal data modeling, querying, and knowledge discovery~\cite{ladner2012mining}. This literature supplies the representation and data-management context for our use of native hydraulic attributes and historical response records.

\subsection{Time-Series Foundation Models}
Chronos-2 uses group attention to share information among related series and supports univariate, multivariate, and covariate-informed forecasting~\cite{chronos2}. We apply its pretrained weights to a source-fitted representation of WSE increments. The frozen baseline provides a common temporal predictor for evaluating the complete project-conditioned retrieval and correction layer.

\subsection{Knowledge and Retrieval Augmentation}
Ioup \emph{et al.} study aggregate $k$-nearest-neighbor queries in spatial networks using the M-tree~\cite{ioup2007aknn}. Their work provides a spatial-network retrieval precedent. Our numerical analog search uses Euclidean distance between reach-local and graph-neighborhood descriptors.

Retrieval-augmented generation combines parametric prediction with external evidence~\cite{rag}. RAF retrieves time-series examples for forecasting~\cite{raf}; TimeRAF learns a retriever and channel-based integration of historical evidence~\cite{timeraf}. Rangaraj \emph{et al.} use similarity and mutual-information retrieval to augment Chronos for daily water-level prediction at Everglades stations~\cite{waterraf2026}.

We investigate project-structured retrieval for hourly HEC-RAS cross-section fields. Native reach relationships, hydraulic attributes, and input mappings determine the retrieval descriptors and residual-correction features. The benchmark compares this coupled method with its frozen temporal base and with classical, recurrent, spectral, and graph-recurrent alternatives.

\section{Approach}\label{sec:approach}
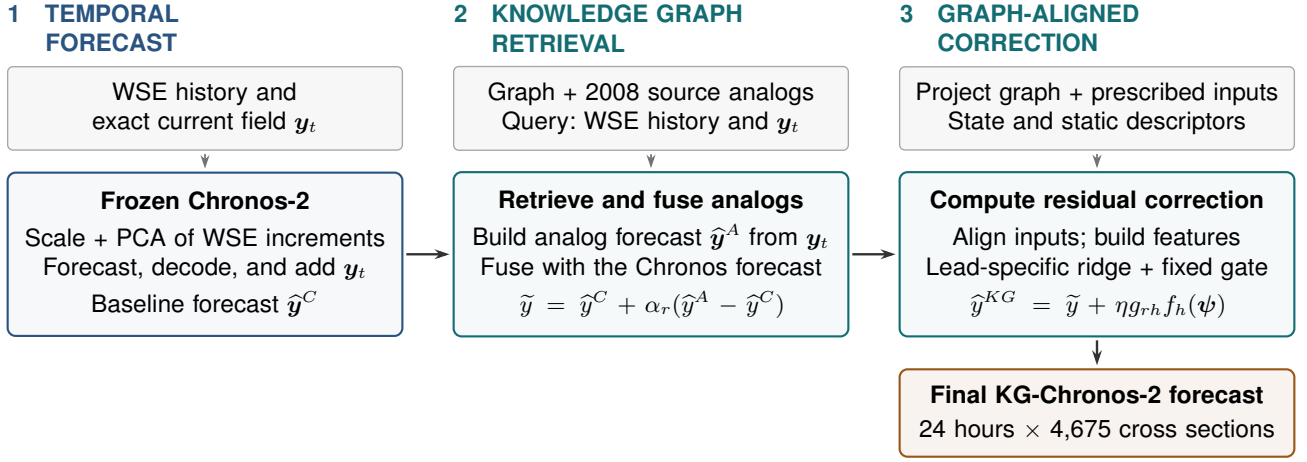
\begin{figure*}[t]
\centering
\input{figures/kg_pipeline}
\caption{KG-Chronos-2 forecast path. Frozen Chronos-2 generates an exact-state baseline. The hydraulic knowledge graph conditions historical-analog retrieval and fusion, then maps prescribed inputs to reaches for gated residual correction. Both the baseline and analog forecast start from $\bm y_t$. Each forecast contains 24 hourly WSE fields at all 4,675 cross sections.}
\label{fig:pipeline}
\end{figure*}

KG-Chronos-2 combines three operations: forecast WSE increments with frozen Chronos-2, fuse the decoded trajectory with graph-conditioned historical analogs, and apply an input-aligned residual correction. Figure~\ref{fig:pipeline} shows the forecast path. All fitted quantities and historical analogs come from the 2008 development simulation. At inference, the model uses the available origin history and prescribed project inputs to produce the complete forecast.

\subsection{Frozen Chronos-2 and Exact-State Forecasting}
We represent hourly WSE increments as $\bm d_t=\bm y_t-\bm y_{t-1}$. Source-training increment standard deviations define a diagonal scale matrix $D$, with safe scales for near-constant channels. Incremental principal-component analysis (PCA) on scaled source-training increments provides mean $\bm\mu$ and loading matrix $C\in\R^{N\times K}$, where $K=256$:
\begin{equation}
\bm z_t=C^{\mathsf T}(D^{-1}\bm d_t-\bm\mu).
\label{eq:encode}
\end{equation}
For each origin, we submit one multivariate task with 512 increments in 256 latent channels to Chronos-2. We disable cross-task learning and request 24 future steps at quantiles 0.1, 0.5, and 0.9. The decoder transforms the latent marginal medians into a native-field point forecast:
\begin{align}
\widehat{\bm d}_{t+h}&=D(C\widehat{\bm z}_{t+h}+\bm\mu),\\
\widehat{\bm y}^{C}_{t+h}&=\bm y_t+\sum_{j=1}^{h}\widehat{\bm d}_{t+j}.
\label{eq:decode}
\end{align}
We select a global increment multiplier on source validation from 0 to 1.5 in steps of 0.05; the selected value is one. The exact origin field bypasses PCA, preserving the full current state during reconstruction.

The baseline uses state history as its predictive input. We fit scaling, PCA, and the multiplier locally while retaining the pretrained Chronos-2 weights. 

\subsection{Hydraulic Knowledge Graph}
Cross-section identities assign the WSE field to 71 reaches. We connect consecutive same-river reaches in native project order, producing 29 undirected pairs that we store as 58 directed adjacency entries before self-loops. Station ordering supplies an approximate orientation for upstream and downstream descriptors.

Fourteen reach descriptors summarize cross-section counts, widths, thalweg and bank statistics, roughness, station extent, structures, and a gradient proxy. We impute and standardize these descriptors across the shared geometry. Hydraulic similarity then reweights the existing adjacency:
\begin{equation}
A^{\mathrm{hyd}}_{rs}=A_{rs}\max\left\{10^{-3},
\exp\!\left(-\frac{\|\bm s_r-\bm s_s\|_2^2}{2\sigma^2}\right)\right\},
\end{equation}
where $\sigma$ denotes the median distance between adjacent reach descriptors. We add self-loops and normalize rows for neighborhood aggregation.

We extract seven input-value channels from HDF \emph{Event Conditions}, match their normalized native identifiers across events, and map them to reaches through river/reach metadata. Broader project-level mappings handle channels with less specific location information. These seven channels define the common input catalog for this benchmark. Table~\ref{tab:knowledge} summarizes the records and their forecasting roles.

\begin{table}[t]
\caption{Hydraulic knowledge in KG-Chronos-2}
\label{tab:knowledge}
\centering\footnotesize
\begin{tabularx}{\columnwidth}{@{}L{0.27\columnwidth}Y@{}}
\toprule
Knowledge object & Link and forecasting role \\
\midrule
Cross-section catalog & Identifier $\rightarrow$ reach membership; aligns outputs and localizes analog reconstruction. \\
Reach graph & Same-river order $\rightarrow$ approximate adjacency; supplies directional descriptors and distance weights. \\
Static attributes & Reach $\rightarrow$ 14 descriptors; reweights adjacency and conditions correction. \\
Input catalog & Native input identifier $\rightarrow$ receiving reaches; aligns seven channels across events. \\
Historical bank & Source event/origin $\rightarrow$ response trajectory; supplies training-segment analogs. \\
\bottomrule
\end{tabularx}
\end{table}

\subsection{Graph-Conditioned Historical Retrieval}
We construct a bank of 2,353 source-training origins at a two-hour stride. Each origin's history and complete 24-hour response fit inside the training segment. Eight state descriptors characterize each query reach: accumulated mean WSE increments over 6, 24, 72, and 168 hours; increment RMS over 24 and 72 hours; and current spatial mean and spread. We combine local, upstream, and downstream aggregates with past-input descriptors, standardize each block using source data, and normalize blocks by dimension.

For each reach, we retrieve the eight nearest source descriptors by Euclidean distance. Inverse-distance weights combine their future increment sequences. The ratio of query to analog 24-hour activity scales each analog, with clipping to $[0.5,2]$. Integrating the weighted increments from the exact query state gives trajectory $\widehat{\bm y}^{A}$. We fit a coefficient $\alpha_r\in[0,1.5]$ on source validation to minimize squared error in the reach-specific fusion:
\begin{equation}
\widetilde y_{t+h,i}=\widehat y^{C}_{t+h,i}
+\alpha_{r(i)}\left(\widehat y^{A}_{t+h,i}-\widehat y^{C}_{t+h,i}\right).
\label{eq:fusion}
\end{equation}
At $\alpha_r=0$, the reach retains the frozen Chronos forecast. Values above one extrapolate along the analog correction.

\subsection{Input-Aligned Residual Correction}
Let $x_{c,j}$ and $v_{r,j}$ denote input increments for channel $c$ and mean WSE increments for reach $r$, respectively. We center and scale each sequence over the complete source-training segment. We then select a lag from overlap-normalized cross-products:
\begin{align}
\rho_{cr}(\tau)&=\frac{1}{n-\tau}\sum_{j=0}^{n-\tau-1}x_{c,j}v_{r,j+\tau},\\
\tau_{cr}&=\argmax_{\tau\in\{0,\ldots,168\}}|\rho_{cr}(\tau)|,
\label{eq:lag}
\end{align}
where $n$ denotes the aligned training length. We clip the selected score to $[-1,1]$ and combine its magnitude with mapped reach distance to weight each channel. We reduce weights for low-score channels and use weaker undirected and disconnected fallbacks. Finally, we normalize channel weights within each receiving reach. The lag supplies an empirical response-alignment parameter.

At lead $h$, the aligned input index is $t+h-\tau_{cr}$. Past features clip this index at the origin. Future features measure the difference between the aligned scheduled input and its clipped past counterpart. The adapter uses 35 features: six past-input, five future-input, eight state, 14 static, and two baseline-forecast descriptors. The last two describe the retrieval-fused mean change and its difference from an auxiliary state-only graph-retrieval forecast.

A separate ridge model $f_h$ predicts the remaining mean reach error at each lead from feature vector $\bm\psi_{t,r,h}$. We broadcast its correction across that reach's cross sections. A source-calibrated gate weights the correction using out-of-fold squared-error improvement:
\begin{equation}
g_{rh}=\clip_{[0,1]}\left[
\frac{1-\SSE^{\mathrm{corrected}}_{rh}/\SSE^{\mathrm{base}}_{rh}}{0.02}
\right].
\label{eq:gate}
\end{equation}
The corrected SSE includes the candidate scale $\eta$. We assign a zero gate when the baseline SSE is at or below $10^{-12}$. The final prediction follows:
\begin{equation}
\widehat y^{KG}_{t+h,i}=\widetilde y_{t+h,i}
+\eta\,g_{r(i)h}\,f_h(\bm\psi_{t,r(i),h}),
\label{eq:kg}
\end{equation}
with selected scale $\eta=0.75$. The gate assigns nonzero weights to 27.35\% of reach--lead pairs. A zero gate retains the retrieval-fused forecast. The fixed gate is retained from source calibration and applied without modification during evaluation.

\section{Methodology and Experimental Setup}\label{sec:setup}
\subsection{Benchmark Design}
Table~\ref{tab:methods} defines the six completed systems. Persistence measures the benefit of forecasting hydraulic change. The residual LSTM tests task-trained temporal prediction using the shared latent representation. Recurrent GeoFNO supplies a spectral and recurrent reference from a recently published hydraulic-surrogate family~\cite{holmberg2025}, and the hydraulic DCRNN-style model tests direct graph-recurrent learning. Frozen Chronos-2 provides the same temporal base as KG-Chronos-2. Their comparison measures the combined addition of project inputs, structured retrieval, and residual correction.

We use the same targets and windows for every system. The three project-conditioned methods receive the same seven-channel catalog through their respective feature representations. We fit LSTM and DCRNN-style candidates with seeds 42, 43, and 44, select one configuration per family by mean source-validation RMSE, and report all three selected fits. The recurrent GeoFNO comparison uses one fresh reference replay.

\begin{table*}[t]
\caption{Six-system benchmark. Every system predicts 24 hourly WSE fields at all 4,675 cross sections. The recurrent GeoFNO and DCRNN-style models receive reach-level input summaries; KG-Chronos-2 retains channel-specific input alignment.}
\label{tab:methods}
\centering\small
\begin{tabularx}{\textwidth}{@{}L{0.19\textwidth}L{0.20\textwidth}L{0.27\textwidth}Y@{}}
\toprule
System & Benchmark role & Forecast-time information & Fitting and generation \\
\midrule
Persistence & Constant-state reference. & Exact current WSE. & Repeat the field at each lead. \\
\addlinespace
Residual LSTM & Task-trained temporal model. & Latent WSE increments and exact current field. & Fit recurrent encoder and decoder; decode 24 predicted increments. \\
\addlinespace
Recurrent GeoFNO & Published-method hydraulic reference. & WSE history, coordinate, static attributes, and prescribed input summaries. & Fit GRU/spectral blocks; recursively advance 24 hourly states. \\
\addlinespace
Hydraulic DCRNN-style & Graph-recurrent hydraulic model. & WSE history, cross-section graph, static attributes, and prescribed input summaries. & Fit diffusion-GRU encoder and decoder; roll out native-field increments. \\
\addlinespace
Frozen Chronos-2 & Pretrained temporal reference. & WSE increment history and exact current field. & Fit scaling/PCA and an increment multiplier; retain pretrained weights. \\
\addlinespace
KG-Chronos-2 & Proposed knowledge-coupled surrogate. & WSE history, reach graph, attributes, source analogs, and prescribed inputs. & Retain Chronos; fit retrieval fusion, response alignment, and gated correction. \\
\bottomrule
\end{tabularx}
\end{table*}

\subsection{Dataset and Forecast Windows}
We use hourly WSE from the native \nolinkurl{MVM_MVK_MVN_Combine} HEC-RAS project. The three arrays share 4,675 cross sections, 71 reaches, and one geometry (Table~\ref{tab:events}). Together they contain 122,424,225 WSE values. The handoff manifest records HEC-RAS version hints of 5.0.1, matching geometry and spatial-order hashes, and successful HDF verification.

\begin{table}[t]
\caption{Dataset and common forecast protocol}
\label{tab:events}
\centering\footnotesize
\begin{tabularx}{\columnwidth}{@{}lllY@{}}
\toprule
Plan & Year & Frames & Role \\
\midrule
p07 & 2008 & 8,737 & Source training, validation, and chronological source test \\
p06 & 2011 & 8,737 & 32 cross-event forecast windows \\
p08 & 2002 & 8,713 & 32 cross-event forecast windows \\
\midrule
\multicolumn{4}{@{}l@{}}{Cadence: 1 hour; domain: 4,675 cross sections, 71 reaches}\\
\multicolumn{4}{@{}l@{}}{Nominal history: 512 hours; forecast: 24 hourly WSE fields}\\
\bottomrule
\end{tabularx}
\end{table}

We partition 2008 at zero-based frame indices $[0,5242)$ for training, $[5242,6989)$ for validation, and $[6989,8737)$ for source testing. Sixteen validation windows support checkpoint and parameter selection. The primary evaluation uses 32 windows from each of 2011 and 2002. Candidate first-target indices begin at 513 with a 24-hour stride; we retain 32 approximately equally spaced candidates per event. This length-based rule gives nonoverlapping target windows and 7,180,800 scalar evaluation values.

The learned models consume a nominal 512-hour history. Chronos and the LSTM use 512 latent increments, which require the preceding WSE level. GeoFNO and the graph-recurrent model encode 512 level/increment pairs. Persistence uses the current field. Timestamps align records and define windows. The project-conditioned models also receive prescribed input schedules over the 24-hour horizon. Future solver outputs serve as evaluation references.

\subsection{Classical and Learned References}
\subsubsection{Persistence}
Persistence carries the exact origin field forward:
\begin{equation}
\widehat{\bm y}^{P}_{t+h}=\bm y_t,\qquad h=1,\ldots,H.
\label{eq:persistence}
\end{equation}
Its error measures the cost of a constant-state forecast.

\subsubsection{Residual LSTM}
The LSTM uses the rank-256 increment representation and exact-state decoder in (\ref{eq:encode})--(\ref{eq:decode}). We further center and scale the latent channels with source-training statistics. A recurrent encoder processes the complete history. A separate recurrent decoder initializes from its hidden and cell states, receives the last observed latent increment, and feeds back each predicted latent increment over the 24-step horizon. A linear readout maps hidden states to 256 output channels. We undo latent normalization, decode native increments, and integrate them from the exact origin field. A source-validation multiplier scales the complete predicted change trajectory.

\subsubsection{Project-conditioned recurrent GeoFNO}
We train a comparator based on a recently published recurrent hydraulic-surrogate architecture~\cite{holmberg2025}. Each cross section supplies standardized WSE levels and increments. A normalized within-reach index provides its coordinate, and 14 static reach descriptors supply hydraulic attributes. Four summaries represent the matched project inputs: the mean and root-mean-square (RMS) of levels and increments. The model receives these summaries over the history and at each forecast step.

A linear encoder and one-layer GRU construct a temporal hidden state. Two residual spectral blocks mix information along the ordered cross sections of each reach. A linear decoder predicts the next-hour increment:
\begin{align}
\widehat{\bm d}_{t+h,r}
 &= f_\theta(\bm H_{h-1,r},\bm s_r,\bm q_{t+h,r}),\label{eq:operator}\\
\widehat{\bm y}_{t+h,r}
 &= \widehat{\bm y}_{t+h-1,r}+\widehat{\bm d}_{t+h,r},
\end{align}
where $\bm H$ denotes recurrent state, $\bm s_r$ contains static descriptors, and $\bm q$ contains input summaries. The rollout starts from the exact origin field and feeds predicted levels and increments back into the shared model for all 24 steps.

\subsubsection{Hydraulic DCRNN-style model}
The graph-recurrent comparator adapts diffusion-GRU encoder--decoder forecasting~\cite{li2018dcrnn} to one node per original cross section. Within each reach, we link consecutive sections in decreasing station order. Approximate upstream reach relationships link reach endpoints. We add self-loops and row-normalize forward and reverse supports. Diffusion convolutions concatenate the input with successive forward and reverse message-passing states and use the result in the GRU gates and candidate update.

Each node receives standardized WSE level and increment, four project-input summaries, its within-reach coordinate, and 14 reach attributes. A recurrent encoder consumes the history. A separate diffusion-GRU decoder uses scheduled inputs and predicted states to generate 24 native-field increments. We integrate these increments from the exact origin and apply a source-selected trajectory multiplier. This hydraulic adaptation uses full cross-section targets and free-running multistep training.

\subsection{Training and Source Selection}
\textbf{Recurrent GeoFNO.} The model uses 48 hidden channels, two spectral blocks, and up to 24 Fourier modes per reach. Its one-step standardized-increment MSE includes a $10^{-5}$ spatial-curvature penalty. We sample reaches in proportion to their cross-section counts and average location losses within each sampled reach. AdamW uses learning rate $2\times10^{-4}$, weight decay $10^{-5}$, and gradient clipping at one. Training allows 20 epochs of 192 minibatches, with batch size at most six and smaller batches for longer reaches. Full 24-step validation occurs at epoch 1 and every second epoch; four checks without improvement stop training. The architecture contains 130,945 parameters.

\textbf{LSTM and DCRNN-style.} We compare LSTM configurations with 128 hidden units/one layer and 256 hidden units/two layers. Graph-recurrent candidates use 16 hidden units/one diffusion step and 32 hidden units/two steps. Both families minimize standardized-increment MSE plus 0.25 times the MSE of lead-normalized cumulative increment error over the free-running 24-step rollout. AdamW uses the same learning rate, weight decay, and clipping values as above. Each seed allows 30 epochs: 64 minibatches of size 16 for the LSTM and 16 minibatches of size one for the graph model. The graph encoder consumes all 512 historical steps and propagates training gradients through the final 64 steps.

After each epoch, we validate complete native-field forecasts and select a trajectory multiplier from 0 to 1.5 in steps of 0.05. A plateau scheduler uses patience two and factor 0.5 on unscaled validation RMSE; five checks without checkpoint improvement stop training. The minimum mean validation RMSE over seeds 42--44 selects the family configuration. Selection retains the two-layer, 256-unit LSTM and the 32-unit, two-step graph model. The reported metrics average their individual-seed forecast errors.

\textbf{KG-Chronos-2.} Alternating source-validation windows calibrate fusion and select an input-descriptor weight from $\{0.25,0.5,1,2\}$. We select two and refit fusion on all 16 windows. Four contiguous source-validation folds select the adapter's ridge penalty from $\{0.01,0.1,1,10,100\}$ and correction scale from $\{0.25,0.5,0.75,1\}$. We refit retrieval-fusion coefficients outside each fold, select penalty ten and scale 0.75, and refit on all validation windows. Section~\ref{sec:discussion} describes shared upstream calibration quantities.

\subsection{Metrics and Paired Comparisons}
For event $e$ and fitted run $s$, we pool squared errors over $W_e$ windows, $H$ leads, and $N$ cross sections. We then average event RMSEs equally:
\begin{equation}
\begin{aligned}
R_{e,s}&=\sqrt{\frac{1}{W_eHN}\sum_{w,h,i}(\widehat y_{s,ewhi}-y_{ewhi})^2},\\
R_{\mathrm{EB},s}&=\frac{1}{E}\sum_{e=1}^{E}R_{e,s},\qquad
\overline R_{\mathrm{EB}}=\frac{1}{S}\sum_{s=1}^{S}R_{\mathrm{EB},s}.
\end{aligned}
\label{eq:rmse}
\end{equation}
Here, $E=2$. For LSTM and DCRNN-style, $S=3$ and we also report the sample standard deviation across fitted runs. Other systems contribute one forecast record. We score each trajectory separately before averaging errors. Relative reduction against reference $b$ is $100(1-\overline R_{\mathrm{EB}}/\overline R_{\mathrm{EB},b})$. We retain native WSE units; the experimental record leaves the physical-unit conversion and datum unresolved.

MAE and final-lead RMSE complement the overall measure. The active-window score uses pre-origin 24-hour increment RMS, aggregated by RMS across reaches. We select its upper quartile within each event, giving eight active windows per event. Active RMSE pools the errors over all locations and leads in those windows; final-lead RMSE uses lead 24 across all 64 windows.

The direct KG-versus-Chronos analysis resamples events and then paired windows within each selected event for 10,000 replicates. Each replicate computes either the event-balanced RMSE difference or the mean of event-wise mean window-RMSE differences. The former averages pooled event RMSEs; the latter averages paired window RMSEs. Percentile bounds at 2.5\% and 97.5\% define the intervals. We keep the fitted forecasts fixed during resampling and report training-seed variability separately. The reported intervals are conditional on this event/window procedure.

\subsection{Execution and Reproducibility}
The saved run uses an NVIDIA RTX PRO 6000 Blackwell Server Edition GPU, PyTorch 2.11.0+cu128, and \texttt{chronos-forecasting==2.3.1}. The reference replay reconstructs the preprocessing and forecasts from the original source pipeline, then the extension fits and evaluates the new baselines. We use its fresh GeoFNO result throughout the expanded comparison. The planned Moirai run stopped during environment creation, before pretrained inference. We report six completed systems and preserve that setup failure in the execution record.

The manuscript package contains saved output evidence, summary tables, source-code excerpts, and figure code. Independent reproduction also requires the simulation HDFs, complete predictions, fitted objects, source-selected checkpoints, exact Chronos weight revision, and documented data access. The immutable source and handoff hashes link the present results to the recorded execution.

\section{Results}\label{sec:results}
\subsection{Performance across Surrogate Approaches}
\begin{figure}[t]
\centering
\includegraphics[width=\columnwidth]{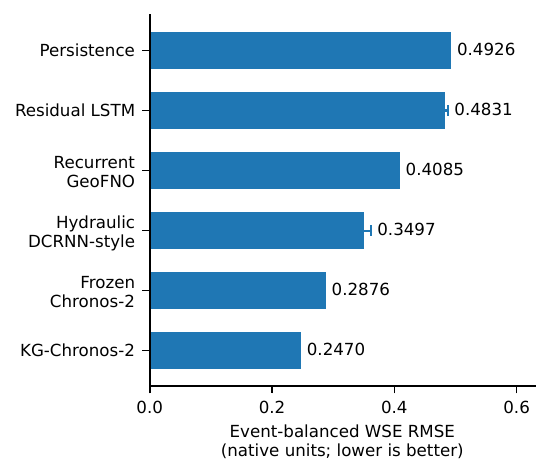}
\caption{Event-balanced RMSE for six completed systems. Whiskers show the sample standard deviation across three training seeds for LSTM and DCRNN-style. The other systems each supply one reference forecast.}
\label{fig:rmse}
\end{figure}

KG-Chronos-2 achieves the lowest event-balanced RMSE among the six completed systems: 0.246970 (Table~\ref{tab:results}, Fig.~\ref{fig:rmse}). It reduces RMSE by 29.38\% relative to the hydraulic DCRNN-style model, 39.54\% relative to recurrent GeoFNO, and 49.87\% relative to persistence. These percentages compare the reported single-run or three-seed mean metrics.

\begin{table*}[t]
\caption{Six-system results in native WSE units. EB averages event-level metrics equally. LSTM and DCRNN-style entries average three individual-seed metrics; $\pm$ gives the sample RMSE standard deviation. Other rows use one forecast record. Bold marks the lowest mean error.}
\label{tab:results}
\centering\small
\setlength{\tabcolsep}{6pt}
\begin{tabular}{@{}lrrrr@{}}
\toprule
System & EB RMSE & EB MAE & EB active RMSE & EB 24-h RMSE \\
\midrule
Persistence & 0.492611 & 0.229447 & 0.654633 & 0.789602 \\
Residual LSTM & 0.483055 $\pm$ 0.005266 & 0.234526 & 0.637714 & 0.778346 \\
Recurrent GeoFNO & 0.408510 & 0.138732 & 0.586309 & 0.656622 \\
Hydraulic DCRNN-style & 0.349729 $\pm$ 0.012500 & 0.119977 & 0.470438 & 0.558798 \\
Frozen Chronos-2 & 0.287613 & 0.085292 & 0.431725 & 0.493318 \\
KG-Chronos-2 & \textbf{0.246970} & \textbf{0.082985} & \textbf{0.351510} & \textbf{0.434783} \\
\bottomrule
\end{tabular}
\end{table*}

The hydraulic DCRNN-style model achieves mean RMSE 0.349729, the lowest among the locally trained neural comparators. Its seed standard deviation is 0.012500. Frozen Chronos-2 reaches 0.287613, reducing the graph-recurrent mean error by 17.76\%. Recurrent GeoFNO reaches 0.408510, a 17.07\% reduction relative to persistence. The residual LSTM reaches $0.483055\pm0.005266$, improving persistence RMSE by 1.94\% while increasing MAE from 0.229447 to 0.234526.

\subsection{Direct Benefit of Knowledge Coupling}
KG-Chronos-2 reduces frozen Chronos-2 RMSE from 0.287613 to 0.246970, a 14.13\% improvement. Its event-balanced MAE decreases from 0.085292 to 0.082985. This contrast evaluates the complete project-input, graph-conditioned retrieval, and residual-correction layer around the same frozen temporal predictor.

\begin{figure}[t]
\centering
\includegraphics[width=\columnwidth]{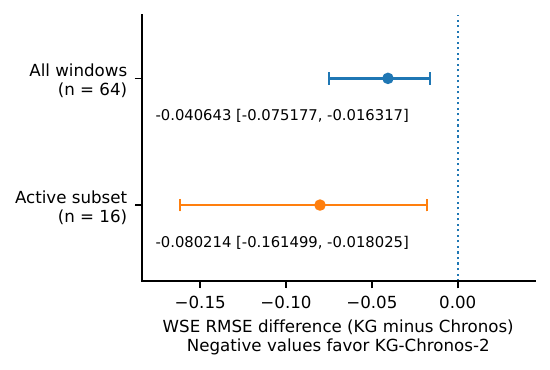}
\caption{Direct KG-Chronos-2 minus frozen Chronos-2 event-balanced RMSE differences. Points and 95\% intervals come from the saved 10,000-replicate paired event/window bootstrap. Negative differences favor KG-Chronos-2.}
\label{fig:paired}
\end{figure}

The paired event-balanced RMSE difference, KG-Chronos-2 minus frozen Chronos-2, is $-0.040643$. Its 95\% hierarchical-bootstrap interval is $[-0.075177,-0.016317]$ from 10,000 replicates (Fig.~\ref{fig:paired}). The mean paired window-RMSE difference is $-0.031776$, with interval $[-0.058094,-0.012728]$. Both intervals favor the coupled system under the recorded resampling procedure. The saved paired summaries also show lower KG-Chronos error in each of the two evaluation events.

\subsection{Active Periods and Final-Lead Error}
\begin{figure}[t]
\centering
\includegraphics[width=\columnwidth]{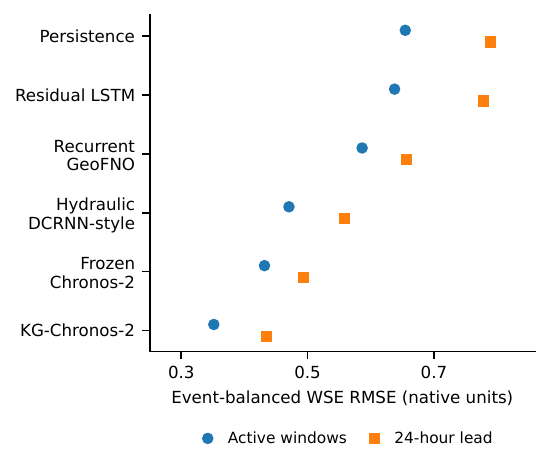}
\caption{Active-window and final-lead RMSE for the same six systems. Active RMSE uses 16 windows selected from pre-origin increments; final-lead RMSE uses lead 24 over all 64 windows. LSTM and DCRNN-style markers average individual-seed metrics.}
\label{fig:dynamic}
\end{figure}

KG-Chronos-2 achieves active-window RMSE 0.351510, compared with 0.431725 for frozen Chronos-2 and a three-seed mean of 0.470438 for the graph-recurrent model (Fig.~\ref{fig:dynamic}). The reduction relative to frozen Chronos is 18.58\%. The saved active event-balanced difference is $-0.080214$, with interval $[-0.161499,-0.018025]$. The mean paired active-window difference is $-0.070229$, with interval $[-0.151480,-0.012435]$.

At lead 24, KG-Chronos-2 reaches RMSE 0.434783, compared with 0.493318 for frozen Chronos-2, 0.558798 for DCRNN-style, 0.656622 for recurrent GeoFNO, 0.778346 for the LSTM, and 0.789602 for persistence. KG-Chronos-2 reduces frozen Chronos error by 11.87\% and recurrent GeoFNO error by 33.78\% at this lead. It achieves the lowest mean error in both evaluation summaries.

\subsection{Variability and Metric Tradeoffs}
The LSTM and graph-recurrent standard deviations describe training-seed variation; the paired KG-versus-Chronos intervals describe event/window resampling for fixed forecasts. KG-Chronos-2 leads the four aggregate measures in Table~\ref{tab:results}, while the LSTM improves on persistence in RMSE and increases MAE.

The original reference evaluation records mean per-event maximum absolute errors of 18.20 for KG-Chronos-2 and 17.49 for persistence, in native WSE units. These averages of two event maxima document an extreme-error tradeoff alongside the aggregate RMSE gain.

\section{Discussion}\label{sec:discussion}
\subsection{Knowledge-Coupled Surrogate Forecasting}
The hydraulic DCRNN-style model provides the strongest locally trained reference by mean RMSE. Frozen \mbox{Chronos-2} improves on that graph-recurrent mean by 17.76\%, and the complete knowledge-coupling layer reduces the frozen model's RMSE by a further 14.13\%. These results support a hydraulic-surrogate design that reuses a pretrained temporal predictor and concentrates project-specific learning in the state representation and knowledge-coupling layer. The direct paired interval supports the augmentation result on the fixed benchmark.

KG-Chronos-2 separates temporal prediction from project-specific adaptation. The frozen predictor forecasts latent increments. Reach relationships organize comparable historical evidence, while input mappings align prescribed disturbances with residual features. The measured gain belongs to their combined forecasting system. Comparisons with matched inputs and correction capacity would quantify the individual contribution of relational structure.

The graph-recurrent comparator learns diffusion-based updates on the cross-section graph; KG-Chronos uses a reach graph to condition retrieval and correction around pretrained forecasts. GeoFNO and DCRNN-style receive aggregate input summaries, while KG-Chronos preserves channel-specific response alignment. Chronos contributes external pretraining, the LSTM and graph model use multistep objectives, and GeoFNO uses a one-step objective with rollout validation. These differences define the selected systems whose errors we compare.

HEC-RAS supplies the development trajectories and project records. Given an exact state history and prescribed inputs, the fitted surrogate computes the next 24 WSE fields directly. This forecast path bypasses numerical time stepping over that horizon. A matched continuation-runtime study would quantify its computational savings.

\subsection{Evaluation Conditions and Further Validation}
We fit task-specific parameters on 2008 and revisited 2002 and 2011 during development. The results describe a fixed, reused, same-geometry benchmark. Two events, overlapping context histories, repeated design choices, and unadjusted exploratory comparisons constrain statistical inference. The paired intervals characterize the recorded forecasts under the stated resampling scheme. An untouched event or project would strengthen transfer evidence.

Source calibration shares upstream fitted quantities across folds. The auxiliary graph-forecast descriptor uses all validation windows, input-feature weighting precedes the adapter folds, and upstream out-of-fold forecasts serve again in second-stage folds. These dependencies can make gate calibration optimistic. Fully nested calibration would rebuild supervised upstream features and select the gate within each outer fold.

The benchmark assumes exact full-domain WSE history and prescribed input schedules. Reach and station ordering approximate connectivity, and seven matched channels define input coverage. Further evaluation should assess missing states, uncertain inputs, verified junction topology, and new geometries. Physical-unit and datum verification would support interpretation of absolute errors. Conservation and local extreme-error diagnostics would assess hydraulic reliability alongside the aggregate scores.

\section{Conclusion}\label{sec:conclusion}
We presented KG-Chronos-2 for 24-hour warm-start forecasting of HEC-RAS WSE. The surrogate combines exact-state residual decoding, a frozen temporal predictor, graph-conditioned historical retrieval, and gated input-aligned correction. It generates forecast fields directly from an available state history and project records.

Across 64 windows at 4,675 cross sections in 71 reaches, KG-Chronos-2 achieves event-balanced RMSE 0.246970 in native WSE units, the lowest among six completed systems. It reduces frozen Chronos-2 error by 14.13\%, with a direct 95\% bootstrap interval favoring the coupled system. Its RMSE is 29.38\% below the three-seed mean of the hydraulic DCRNN-style model and 39.54\% below the replayed recurrent GeoFNO. It also achieves the lowest active-window and final-lead RMSE. The study demonstrates the forecasting value of coupling a frozen temporal model to native hydraulic project knowledge on the shared-geometry benchmark.

\IEEEtriggeratref{17}
\bibliographystyle{IEEEtran}
\bibliography{references}
\end{document}

%% file: heading_punctuation.tex
\makeatletter
\renewcommand{\abstract}{\normalfont
  \@IEEEabskeysecsize\bfseries\textit{\abstractname}:\enspace
  \@IEEEgobbleleadPARNLSP}
\renewcommand{\IEEEkeywords}{\normalfont
  \@IEEEabskeysecsize\bfseries\textit{\IEEEkeywordsname}:\enspace
  \@IEEEgobbleleadPARNLSP}
\makeatother

%% file: figures/kg_pipeline.tex
\begingroup%
\renewcommand{\sfdefault}{phv}
\definecolor{kgTemp}{RGB}{42,83,128}%
\definecolor{kgKnow}{RGB}{20,108,115}%
\definecolor{kgOut}{RGB}{152,87,24}%
\begin{tikzpicture}[
  x=1cm,y=1cm,
  font=\sffamily\fontsize{9}{11}\selectfont,
  source/.style={
    draw=black!35,fill=black!3,line width=0.5pt,
    rounded corners=2pt,align=center,
    text width=4.80cm,inner xsep=6pt,inner ysep=6pt,
    minimum height=1.05cm,outer sep=0pt
  },
  process/.style={
    draw=kgTemp,fill=kgTemp!4,line width=0.8pt,
    rounded corners=3pt,align=center,
    text width=4.80cm,inner xsep=6pt,inner ysep=7pt,
    minimum height=2.16cm,outer sep=0pt
  },
  stage/.style={
    anchor=west,inner sep=0pt,align=left,text width=5.22cm,
    font=\sffamily\bfseries\fontsize{9}{11}\selectfont
  },
  flow/.style={-{Stealth[length=2mm,width=1.3mm]},
    draw=black!80,line width=0.9pt,shorten <=1.5pt,shorten >=1.5pt},
  feed/.style={-{Stealth[length=1.8mm,width=1.2mm]},
    draw=black!55,line width=0.65pt,shorten <=1.5pt,shorten >=1.5pt},
  result/.style={
    draw=kgOut,fill=kgOut!7,line width=0.8pt,
    rounded corners=3pt,align=center,
    text width=4.80cm,inner xsep=6pt,inner ysep=6pt,
    minimum height=0.86cm,outer sep=0pt
  }
]
\node[stage,text=kgTemp] at (0,2.99)
  {1\quad TEMPORAL\\\hphantom{1\quad}FORECAST};
\node[stage,text=kgKnow] at (5.90,2.99)
  {2\quad KNOWLEDGE GRAPH\\\hphantom{2\quad}RETRIEVAL};
\node[stage,text=kgKnow] at (11.80,2.99)
  {3\quad GRAPH-ALIGNED\\\hphantom{3\quad}CORRECTION};

\node[source] (history) at (2.61,1.93)
  {WSE history and\\exact current field $\bm y_t$};
\node[source] (knowledge) at (8.51,1.93)
  {Graph + 2008 source analogs\\
   Query: WSE history and $\bm y_t$};
\node[source] (inputs) at (14.41,1.93)
  {Project graph + prescribed inputs\\
   State and static descriptors};

\node[process] (chronos) at (2.61,0)
  {\textbf{Frozen Chronos-2}\\[3pt]
   Scale + PCA of WSE increments\\
   Forecast, decode, and add $\bm y_t$\\[3pt]
   Baseline forecast $\widehat{\bm y}^{C}$};
\node[process,draw=kgKnow,fill=kgKnow!4] (fusion) at (8.51,0)
  {\textbf{Retrieve and fuse analogs}\\[3pt]
   Build analog forecast $\widehat{\bm y}^{A}$ from $\bm y_t$\\
   Fuse with the Chronos forecast\\[3pt]
   $\widetilde y=\widehat y^{C}+\alpha_r(\widehat y^{A}-\widehat y^{C})$};
\node[process,draw=kgKnow,fill=kgKnow!4] (correction) at (14.41,0)
  {\textbf{Compute residual correction}\\[3pt]
   Align inputs; build features\\
   Lead-specific ridge + fixed gate\\[3pt]
   $\widehat y^{KG}=\widetilde y+\eta g_{rh} f_h(\bm\psi)$};

\draw[flow] (chronos.east) -- (fusion.west);
\draw[flow] (fusion.east) -- (correction.west);
\draw[feed] (history.south) -- (chronos.north);
\draw[feed] (knowledge.south) -- (fusion.north);
\draw[feed] (inputs.south) -- (correction.north);

\node[result] (output) at (14.41,-2.10)
  {\textbf{Final KG-Chronos-2 forecast}\\[2pt]
   24 hours $\times$ 4,675 cross sections};
\draw[flow] (correction.south) -- (output.north);
\end{tikzpicture}%
\endgroup%